\documentclass[sigconf]{acmart}
\AtBeginDocument{%
  }
\newcommand{\xmark}{\ding{55}}
\newcommand{\cmark}{\ding{51}}

\usepackage{graphicx}
\usepackage{booktabs}
\usepackage{subcaption}
\usepackage{multirow}
\usepackage{adjustbox}
\usepackage{multirow}
\usepackage{graphicx}
\usepackage{hyperref}
\usepackage{url}
\usepackage{natbib}
\usepackage{booktabs}
\usepackage{subcaption}
\usepackage{footnote}
\usepackage{pifont}
\usepackage{float}
\usepackage{wrapfig}
\usepackage{xcolor}
\usepackage{rotating}
\usepackage{array}
\usepackage{arydshln}

\newcolumntype{C}[1]{>{\centering\arraybackslash}p{#1}}
\setcopyright{acmlicensed}
\copyrightyear{2018}
\acmYear{2018}
\acmDOI{XXXXXXX.XXXXXXX}

\acmConference[DAC]{The Design Automation Conference}{June 22--25,
  2025}{San Francisco, CA}
\acmISBN{978-1-4503-XXXX-X/18/06}

\begin{document}

\title{Logic Synthesis Optimization with Predictive Self-Supervision via Causal Transformers}

\author{Raika Karimi}
\email{raika.karimi@huawei.com}
% \city{Toronto}
% \country{Canada}
\affiliation{%
  \institution{Huawei Noah’s Ark Lab}
  \city{Toronto}
  \country{Canada}
}

\author{Faezeh Faez}
\email{faezeh.faez@huawei.com}
\affiliation{%
  \institution{Huawei Noah’s Ark Lab}
  \city{Toronto}
  \country{Canada}
}

\author[1]{Yingxue Zhang}
\email{yingxue.zhang@huawei.com}
\affiliation{%
  \institution{Huawei Noah’s Ark Lab}
  \city{Toronto}
  \country{Canada}
}

\author{Xing Li}
\email{li.xing2@huawei.com}
\affiliation{%
  \institution{Huawei Noah’s Ark Lab}
  \city{Hong Kong}
  \country{China}
}

\author{Lei Chen}
\email{lc.leichen@huawei.com}
\affiliation{%
  \institution{Huawei Noah’s Ark Lab}
  \city{Hong Kong}
  \country{China}
}

\author{Mingxuan Yuan}
\email{yuan.mingxuan@huawei.com}
\affiliation{%
  \institution{Huawei Noah’s Ark Lab}
  \city{Hong Kong}
  \country{China}
}

\author{Mahdi Biparva}
\email{mahdi.biparva@huawei.com}

\affiliation{%
\institution{Huawei Noah’s Ark Lab}
  \city{Toronto}
  \country{Canada}
}

% \affiliation{\institution{Huawei Noah’s Ark Lab}}
\renewcommand{\shortauthors}{Trovato et al.}


\begin{abstract}
% - What are we doing in this paper? Why is it relevant?
Contemporary hardware design benefits from the abstraction provided by high-level logic gates, streamlining the implementation of logic circuits. 
% This abstraction, known as Logic Synthesis Optimization (LSO) 
% \MB{It is vague and doubtful to say the abstraction is logic synthesis optimization! here a reliable definition of LSO should be given, where lso aims to optimize the logical gates inter-connectivity according to some utility function such as area or delay}, within the Electronic Design Automation (EDA) workflow aims to enhance logic circuits for performance metrics like size and speed in the final layout.
Logic Synthesis Optimization (LSO) operates at one level of abstraction within the Electronic Design Automation (EDA) workflow, targeting improvements in logic circuits with respect to performance metrics such as size and speed in the final layout.
Recent trends in the field show a growing interest in leveraging Machine Learning (ML) for EDA, notably through ML-guided logic synthesis utilizing policy-based Reinforcement Learning (RL) methods.
% - why is it hard? Indicate the significance of our work?
Despite these advancements, existing models face challenges such as overfitting and limited generalization, attributed to constrained public circuits and the expressiveness limitations of graph encoders. 
% - How do we solve it? What is our contribution? List them all.
To address these hurdles, and tackle data scarcity issues, we introduce LSOformer, a novel approach harnessing Autoregressive transformer models and predictive SSL to predict the trajectory of Quality of Results (QoR). 
LSOformer integrates cross-attention modules to merge insights from circuit graphs and optimization sequences, thereby enhancing prediction accuracy for QoR metrics. 
% - How do we verify we solved it? Is it empirical? What is the experimental setup?
Experimental studies validate the effectiveness of LSOformer, showcasing its superior performance over baseline architectures in QoR prediction tasks, where it achieves improvements of 5.74\%, 4.35\%, and 17.06\% on the EPFL, OABCD, and proprietary circuits datasets, respectively, in inductive setup.
% Experimental evaluations underscore the efficacy of LSOformer, showcasing its superior performance over baseline architectures in QoR prediction tasks.

% - Elaborate on the empirical evidence supporting we solve it; main table, ablation studies, analysis
% \MB{a statement giving some insight on what empirically we achieve is helpful; e.g. how much we improve baseline/sota}
\end{abstract}



\received{20 February 2007}
\received[revised]{12 March 2009}
\received[accepted]{5 June 2009}

\maketitle

\section{Introduction}

% - What are we doing in this paper?
In contemporary hardware design, the creation of hardware devices is facilitated by the abstraction provided by high-level logic gates, liberating designers from the intricacies of Integrated Circuit (IC) implementation. During the IC fabrication process, the conceptual designs are translated into tangible layouts with the aid of Electronic Design Automation (EDA) tools. In essence, EDA tools afford designers the opportunity to focus exclusively on the functional aspects of their designs at a high-level by employing hardware description languages (HDL) like Verilog.

% - Why is it relevant?
logic synthesis optimization (LSO) in which a logic circuit is transformed into a functionally equivalent and optimized representation  is a part of EDA workflow \cite{sasao1993logic}. In LSO, a sequence of optimization transformations supported by academia and industry \cite{amaru2017logic} and named \textbf{recipe} are applied to logic-level circuit designs to optimize performance criteria such as size and speed, named Quality of Results (QoR), in the final IC. 

Due to high latency and long running time of EDA tools, there is a huge surge of interest to use Machine Learning (ML) for EDA \cite{yang2022versatile,huang2021machine,ghose2021generalizable,chowdhury2023towards,sohrabizadeh2023robust}, 
% specifically ML-guided logic synthesis in our problem \cite{yang2022logic}. One way Logic Synthesis designers uses is to uncover series of optimization sequences by using policy-based Reinforcement Learning (RL) methods \cite{grosnit2022boils}. These searches can be time consuming; hence, a pre-trained policy functions can be used to accelerate RL methods \cite{chowdhury2024retrieval}. Two aforementioned reasons  motivated the researches to open up a new problem to come up with ML models with high generalization power to predict final QoR given gate-level representation of circuits and optimization sequences \cite{chowdhury2021openabc}. 
specifically ML-guided logic synthesis \cite{huang2021machine}, \cite{yang2022logic}. In addressing this, logic synthesis designers frequently employ policy-based Reinforcement Learning (RL) methods to identify optimal recipes \cite{grosnit2022boils}. 
Given the time-intensive nature of searches, leveraging pre-trained policy functions can expedite RL processes \cite{chowdhury2024retrieval}. 
This has introduced a new research challenge: developing ML models that act as policy functions or serve as alternatives to running costly EDA pipelines, capable of accurately mapping a pair of gate-level circuit representations and a recipe to the final QoR \cite{chowdhury2021openabc}.
% This has motivated researchers and us to design generalized ML models capable of accurately mapping a pair of gate-level circuit representations and a recipe to the final QoR .
% This has led researchers and us to develop ML models with enhanced generalization, leveraging gate-level circuit representations and recipes to accurately predict the final QoR.
% This has led researchers to explore new challenges in developing ML models that exhibit robust generalization capabilities for predicting the final QoR based on gate-level circuit representations and optimization sequences \cite{chowdhury2021openabc}.

% The most of these models inputs logic-level circut designs in AIG form \cite{mishchenko2006dag} to benefit from inductive bias of Directed Acyclic Graph (DAG) structure \cite{neto2019lsoracle}. This track of research is one of the active research areas to inject the inductive bias of DAG into graph encoders \cite{luo2024transformers} which is a useful inductive bias to generate suitable representation for DAGs. 

% Most models in this domain process logic-level circuits in the And-Inverter Graph (\textbf{AIG}) format \cite{mishchenko2006dag} which has Directed Acyclic Graph (DAG) structure characterized by partially ordered nodes \cite{neto2019lsoracle}. Extensive research has focused on incorporating the DAG's inductive bias into message-passing graph encoders \cite{thost2021directed} or Graph Transformers \cite{luo2024transformers} to improve model expressiveness. 
Most models in this domain represent logic-level circuit designs using the And-Inverter Graph (\textbf{AIG}) format \cite{mishchenko2006dag}, which adopts a Directed Acyclic Graph (DAG) structure with partially ordered nodes \cite{neto2019lsoracle}. Significant research efforts have aimed at embedding the inductive bias of DAGs into message-passing graph encoders \cite{thost2021directed} and graph transformers \cite{luo2024transformers} to enhance model expressiveness. One contribution of our work extends these concepts by generating graph-level representations specifically optimized for AIGs.

The current QoR prediction models often experience overfitting due to a limited number of publicly available circuits used during training, resulting in a lack of generalization, especially when tested on unseen circuits that differ in size and characteristics from the training circuits. Additionally, these models struggle with imbalanced input samples across different modalities; typically, there are more recipe available than AIGs. This disparity can lead to prominent overfitting, particularly due to the low expressiveness of graph encoders, which becomes exacerbated when the model excessively adapts to the encoder part handling recipe.

% Recently, self-supervised learning has addressed both the scarcity of data, offering an alternative to costly data augmentation methods, and issues of expressiveness in various downstream tasks related to LSO \cite{chowdhury2023converts, wang2022functionality}. This approach involves pre-training the circuit encoder component in advance, followed by fine-tuning it on specific downstream tasks. Inspired by predictive SSL \cite{liu2022graph}, another contribution of our work is the definition of an auxiliary task that involves predicting intermediate QoRs. This joint training mechanism provides a stronger supervisory signal during training, enhancing the learning process.

Self-supervised learning (SSL) has recently emerged as a powerful framework to mitigate data scarcity and address expressiveness limitations in downstream tasks related to LSO \cite{chowdhury2023converts, wang2022functionality}. By pre-training circuit encoders and fine-tuning them for specific applications, SSL offers an efficient alternative to conventional data augmentation. Inspired by predictive SSL paradigms \cite{liu2022graph}, this work introduces an auxiliary task that predicts intermediate QoRs, enabling a joint training approach that strengthens the supervisory signal and enhances model learning dynamics.

% \MB{incomplete sentence ... by pre-training circuit encoder part a priori and then fine-tuning on downstream tasks}. 
% if we include data augmentation algorithm
% moreover, we have proposed a data augmnetation algorithm to generate bunch of IPs with different scales.
% if we have cross attention worked properly
% Another drawback of existing method is encoder used to encode OS has no information from the input graphs and there is information bottleneck in fusing graph embedding and OS embedding when normal concatenation applied. 
Existing approaches face a critical limitation in the recipe encoder, which operates in isolation from AIG representations, leading to an inherent information bottleneck. This limitation becomes particularly evident when AIG and recipe embeddings are fused using conventional concatenation methods.
% restricting the model's capacity to effectively integrate contextual information.
% Another drawback of existing method is that the encoder used for encoding recipe lacks information from the input graphs, leading to an information bottleneck. This bottleneck occurs particularly when graph embeddings and recipe embeddings are merged using standard concatenation techniques. 
% This motivated us to come up with decoder-only transformer for the training given the joint loss function. Therefore, the model contextualize Heuristics's embedding using AIG embeddings through Cross-Attention module. 
This limitation motivated us to develop a decoder-only transformer model for training, employing a joint loss function. As a result, the model contextualizes the embeddings of heuristics using AIG embeddings through a Cross-Attention module, facilitating more effective integration of information.
% \MB{some context on how the over-fitting/lack of training data is approached in the community, for instance SSL pre-training, data augmentation etc. This is a context for our contribution later.}

% \MB{- How do we solve it? What is our contribution? We need to . \\
% - - inspired by predictive SSL, we define an auxiliary task to provide stronger supervisory signal during the training
% - - we formulate it as join training of the predictive SSL objectives and the downstream task
% - - decoder-only transformer for the training given the joint loss function.}
In summary, we propose the following innovations to address aforementioned drawbacks:
\begin{itemize}
% Level-wise Graph Pooling: An efficient graph pooling mechanism for Directed Acyclic Graph (DAG) structures that transforms And-Inverter Graphs (AIGs) into a sequence of embeddings.
  \item \textbf{Level-wise Graph Pooling:}  An efficient graph pooling mechanism tailored for DAG structures that transforms AIG into a sequence of embeddings.
  % Level-wise graph pooling, An efficient Graph pooling mechansim for DAG structure to convert the AIGs to a sequence of embeddings. 
  
  \item \textbf{QoR Trajectory Prediction:} An auxiliary predictive SSL task designed to predict intermediate QoR for joint training. 
  % An Auxiliary predictive SSL task to predict intermediate QoRs for joint training.
  \item \textbf{Causal Transformer:} A decoder-only transformer model that causally predicts QoR based on the recipes and AIGs.
  % A decoder-only transformer to causally predict QoRs given the OS and AIGs. 
  \item \textbf{Contextualized Fusion:} An efficient mechanism to fuse embeddings of AIGs and recipes using a cross-attention module. This approach contextualizes the embeddings of recipes with AIGs, enabling the model to learn which parts of the graph each heuristic attends to.
  % An effiencint mechanism to fuse embeddigns of AIGs and OSs using cross-attention module to contextulized the embeddings of recipes with AIGs and to let model learng which part of graph each heuristic attends to.

\end{itemize}

% \MB{summary of contributions}

%%%%%%%%%%%%%%%%%%%%%%%%AIG representation.
% We have tested our SOTA model on several datasets and shows its efficiency in terms MAPE and task-specific metrics compared to baseline architecture for QoR prediction. Moreover, we justify with our ablation study that our novel architecutre fro downstream task is highly aligned with predictive SSL task so the combined performance outperfroms individual architecure. this is also another sign of that low-dataset regime issue has been addressed by ssl task. 

Our SOTA model has been tested across multiple datasets, showcasing superior performance in terms of Mean Absolute Percentage Error (MAPE) when compared to conventional baseline architectures for QoR prediction. Furthermore, our ablation studies justifies that the novel architecture designed for downstream task aligns effectively with predictive SSL tasks. This synergy enhances overall performance beyond that of individual models, also suggesting that the challenges associated with limited datasets have been effectively mitigated through the SSL approach.

%%%%%%%%%%%%%%%%%%%%%
% \MB{this part onwards needs to be written again according to the comments provided above.}
% \MB{
% To address all aforementioned drawbacks, here we propose LSOformer, which benefits from high expressiveness power of transformers coupled with predictive SSL pre-text tasks aligned with downstream task, QoR prediction. Moreover, information fusion bottleneck has been addressed by adding cross attention module to let model learn which part of graph each optimization sequence attend to. We have tested our SOTA model on several datasets and shows its efficiency in terms MAPE and task-specific metrics compared to baseline architecture for QoR prediction.
% }

%%%%%%%%%%%%%%%%%%%%%%%%%%Related Work%%%%%%%%%%%%%%%%%%%%%%%%%%%%%
\section{Related Work}

% - Representation Learning for Logic Synthesis Optimization:
% In this part, we need to list all the related papers that propose and present novel approaches to learning-based prediction methods for LSO. We need to compare and contrast. It is good to categorize and cluster them together based on some criteria.
% (this is shared with both of the papers, this and the other one, you can share the list of papers, but the text must be written individually).

Logic synthesis necessitates thorough adjustment of the synthesis optimization procedure, with the QoR contingent upon the optimization sequence applied. Effectively exploring the design space presents a challenge due to the exponential array of potential optimization permutations \cite{feng2022batch}. Consequently, fast and efficient automation of the optimization procedure becomes imperative \cite{li2023effisyn,shi2023deepgate2}. \cite{hosny2020drills, chowdhury2022bulls, chowdhury2024retrieval, grosnit2022boils, qian2024efficient, wang2023easymap} introduce innovative  methodologies based on Qlearning-based, Bayesian Optimization, and policy-based RL, which autonomously traverse the optimization space, obviating the need for human intervention.

Several policy-based RL methodologies, such as those discussed in \cite{yang2022logic, chowdhury2024retrieval, wu2022ai}, leverage a pre-trained function that correlates initial circuits and recipe with the ultimate QoR. This architecture is split into two primary branches: one for encoding AIGs and another for recipes, along with a fusion branch that maps the embeddings onto the predicted QoR. Similar efforts to forecast QoRs based on initial graphs and recipe are documented in \cite{chowdhury2021openabc, yang2022logic, yang2022prediction, wu2022lostin, wu2022ai, zheng2024lstp}, where various graph and sequence encoders are explored.
% uncomment start raika
Discussions abound concerning the most effective AIG encoders and methods for integrating sequence embeddings with graph embeddings. As a result, improvements in encoder and fusion modules are crucial for enhancing prediction performance.
% uncomment end raika
For example, \cite{chowdhury2021openabc} utilizes a GCN to encode graphs and a CNN for recipes, employing both mean and max pooling for graph combination. Meanwhile, \cite{yang2022prediction} employs GraphSage for graph encoding and a transformer for recipe encoding. Similar to one of recent baselines, LOSTIN \cite{wu2022lostin}, \cite{yang2022logic} integrates a Graph Isomorphism Network (GINE) for graph learning with an LSTM for recipe encoding. Additionally, \cite{wu2022ai} explores various graph encoders, such as PNA and pooling techniques, alongside virtual supernodes in conjunction with hierarchical learning and RL methods. In all aforementioned methods, recipe embeddings and Graph embeddings are fused using concatenation. The MLP decoder has been used to generate the predicted QoR after fusion. However, none of these models utilize an attention mechanism to fuse the two branches.

\begin{table}[t]
\centering
\caption{Overall notation table of the main symbols in the paper. 
}\label{tab:notation}
\resizebox{0.8\linewidth}{!}{%
\begin{tabular}{c c}
\hline 
\multicolumn{2}{c}{\textbf{basic notations}}\\ \hline\midrule
$N$ & The number of nodes.\\
$E$ & The number of edges.\\
$D$ & The maximum depth of the AIG\\
$C$ & The number of available heuristics\\
$R$ & The number of available recipes\\
$M$ & The maximum length of recipe\\
% \FF{$F_j$} & \FF{...}\\
$\hat{y}$ & The prediction of final ground truth QoR\\
$y_{k}$ & The ground truth QoR of $k^{th}$ step \\
% $QoR^i_{GT}$ & Ground Truth QOR at $i^{th}$ step\\
% $QoR_{GT}$ & Ground Truth QOR at final step\\
% $QoR_{prediction}^i$ & Predicted Last QOR \\
\hline
\multicolumn{2}{c}{\textbf{Sets}} \\ \hline
\midrule
$\boldsymbol{H}$ & The set of node embeddings  \\
$\boldsymbol{H}^l$ &  The set of embeddings for nodes located at $l^{th}$ level of AIG. \\
$\mathcal{G}$ & AIG Graph \\
$\mathcal{V}$ & The set of vertices.\\
$\mathcal{E}$ & The set of edges    \\
$\mathcal{X}^v$ & The node features.   \\
$\mathcal{X}^v_{type}$ & The node type\\
$\mathcal{X}^v_{inverted}$ &The number of inverted inputs for each node\\
$\mathcal{T}$ & Set of Available Heuristics\\
% $\mathcal{D}$ & The set of node's depth \\
$\mathcal{R}$ & The set of recipes   \\
% $\mathcal{H}$ & The set of Heuristics  \\
$u_{i}$ & $i^{th}$ heuristic within a given recipe \\
$d_{h}$ & the node embedding dimension. \\\hline
\multicolumn{ 2}{c}{\textbf{Sequences}} \\ \hline
\midrule
% $QoR_{trajectory}$ & Sequence of QORs at all steps \\
$\bar{H}$ & Sequence representation of AIG \\
$r_{i}$ & $i^{th}$ recipe \\
$\bar{r}_{i}$ & Sequence of embeddings for $i^{th}$ recipe \\
$z_{i}$ &  Sequence of positionally encoded embeddings for $i^{th}$ recipe \\
$\bar{z}_{i}$ & Contextualized heuristics' embeddings  \\
$\widetilde{h}^{i}$ &   Output sequence of embeddings for input AIG and $i^{th}$ recipe\\\hline
\multicolumn{2}{c}{\textbf{Matrices and Vectors}} \\ \hline
\midrule 
$\bar{u}_{i}$ &  Embedding vector for $i^{th}$ heuristic of a given recipe \\
% $\Tilde{f}_{i}$ & Output Embedding vector for $i^{th}$ step of a given recipe \\
$h_i$ &  $i^{th}$ node embedding  \\
$\bar{M}$ &  Upper triangular mask matrix \\
$\bar{h}_l$ &  The emebdding of $l^{th}$ level of AIG \\\hline
\multicolumn{2}{c}{\textbf{Learnable Parameters and functions}} \\ \hline
\midrule
${\theta}$ &  The weight parameters of the encoder. \\ 
${\gamma}$ &  The weight parameters of the decoder. \\\hline
\bottomrule
\end{tabular}%
}
\end{table}

% \cite{wang2023easymap}
% \cite{shi2023deepgate2}

\textbf{SSL for EDA:}
% Different types of SSL such as predictive and contrastive learning has shown significant improvement on graph domain tasks \cite{wu2021self}. Different SSL protocols can be defined to pre-train or jointly train graph encoders using graph-level pre-text tasks \cite{liu2022graph}.
% Another track that improves the expressiveness power of AIG encoder and mitigate data scarcity is Self-Supervised Learning.   For example, \cite{chowdhury2023converts} utilizes contrastive learning for pre-training AIG encoders to perform Netlist classification task. \cite{wang2022functionality} baseline for Contrastive-learning on netlist representation.
Various forms of SSL, including predictive and contrastive approaches, have demonstrated substantial enhancements in tasks within the graph domain \cite{wu2021self}. Different SSL protocols can be devised for pre-training or jointly training graph encoders through graph-level pretext tasks \cite{liu2022graph}. 
% uncomment start raika
Additionally, SSL provides an alternative to costly data augmentation for AIGs \cite{li2023verilog}, offering a promising method to enhance the expressiveness of AIG encoders and tackle issues related to data scarcity. 
% uncomment end raika
For instance, \cite{chowdhury2023converts} employs contrastive learning to pre-train AIG encoders for the Netlist classification task. Similarly, \cite{wang2022functionality} establishes a baseline for contrastive learning in netlist representation.

% \MB{
% - Predictive (Graph) SSL 
% What is it, what are success cases in the graph domain, loss formulation. You can look into the Graph SSL surveys. Point out how different training protocols are defined in graph ssl, pre-training then fine-tuning, join-training, etc.
% }

% \MB{
% - Causal Transformers for Sequential Data
% In this part, we need to review approaches under Transformers where sequential data are modeled for some predictive task. This includes various attention mechanisms, i.e. self-attention, cross-attention etc
% }

% There is an active research field to use predictive SSL to improve the expressiveness of transformer-based encoders. For instance, \cite{rong2020self} benefits from motif prediction to enrich Graph transformer encoder. In sequence domain, researchers predict the next tokens using causal transformers to leverage predictive SSL \cite{rohekar2024causal} or for next token prediction which boost the performance of pure transformers \cite{vaswani2017attention}.

The application of predictive SSL to enhance the expressiveness of transformer-based encoders constitutes a dynamic area of research. For example, \cite{rong2020self} demonstrates the use of motif prediction to augment the capabilities of graph transformer encoders. In the domain of sequence processing, researchers employ causal transformers with predictive SSL to enhance their expressiveness \cite{rohekar2024causal}. Additionally, next-token prediction exemplifies the use of causal pre-training, which substantially improves the effectiveness of conventional transformers \cite{vaswani2017attention}.

\begin{figure*}[t]
    \centering
    \includegraphics[width=0.99\textwidth,height=7.8cm]{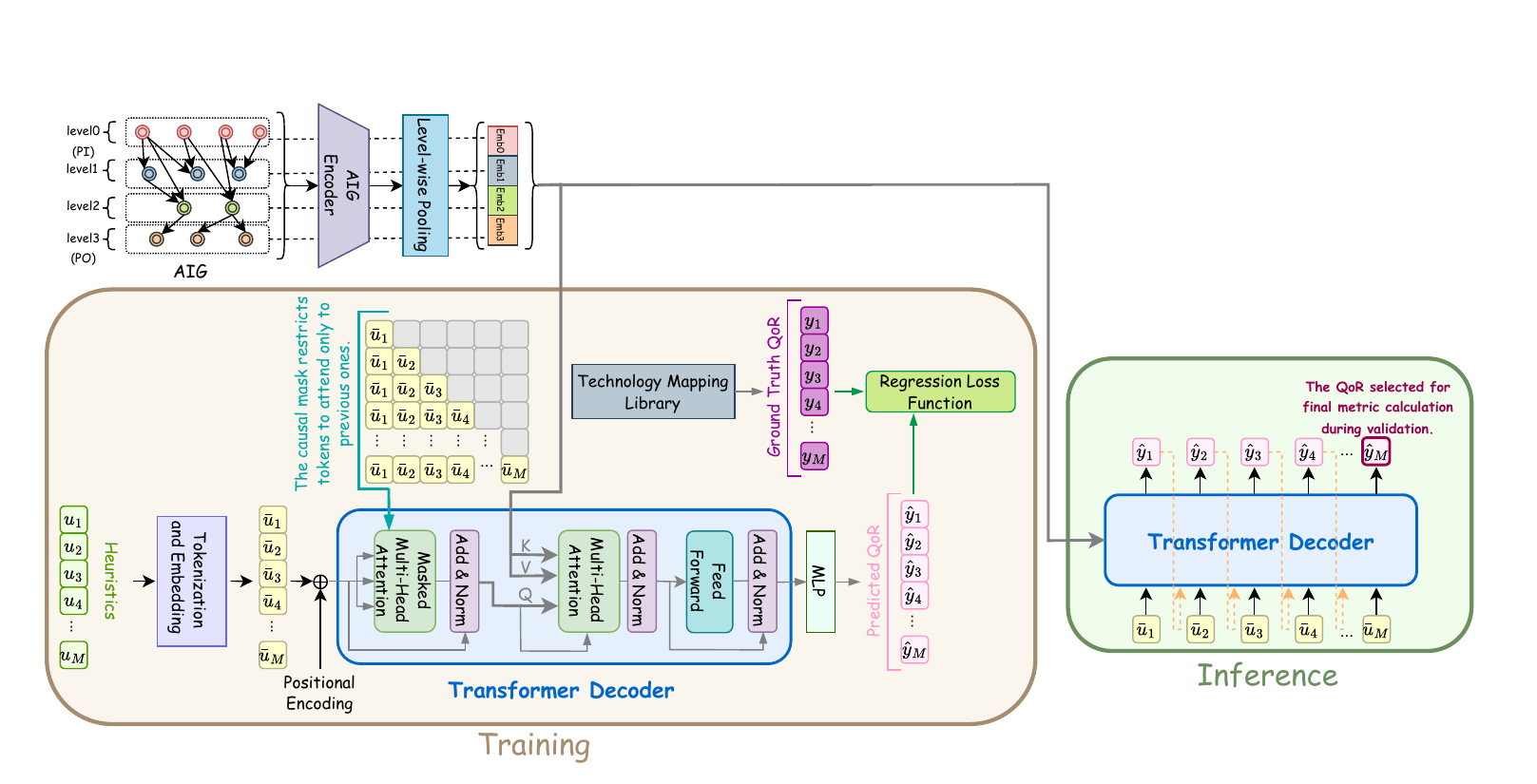}
    \caption{ Schematic overview of the LSOformer architecture during training and inference modes. 
    }
    \label{fig:2}
\end{figure*}

%%%%%%%%%%%%%%%%%%%Methodology%%%%%%%%%%%%%%%%%%%%%%%%%%%%%%%%%%%%%
\section{Methodology}
%%%%%%%%%%%%%%%%%%%%%%%%%%%%%%%%%%%%%%%%%%%%%%%%%%%%%%%%%%%%%%
In this section, we formalize and describe the Logic Synthesis Optimization (LSO) problem. Next, we present the LSOformer pipeline, specifically designed for LSO, which features a novel architecture tailored for the QoR prediction task. Then, we introduce a Self-Supervised Learning (SSL) auxiliary task aiming at predicting the trajectory of the QoRs given recipe and And-Inverter Graphs (AIG) graphs.

% \FF{In this section, we formalize and describe the Logic Synthesis Optimization (LSO) problem. Next, we present the LSOformer pipeline, specifically designed for LSO, which features a novel architecture tailored for the Quality of Result (QoR) prediction task. Additionally, we introduce a self-supervised learning auxiliary task aimed at predicting the trajectory of QoRs based on Optimization Sequence (OS) and AIG graphs.}

% - Problem statement/formulation/preliminaries/background
% We need to introduce the problem setting, notation (formalism) of the work. Additionally, we need to highlight any specific assumptions that are unusual.
% - - we need to define notations and terminologies and stick with them throughout the paper.
% - - the notation is needed to describe the data domain, the input, the target, the downstream regression task, the learning models, joint training (predictive SSL and downstream task)
% \MB{we need notation for the input, the model, and the downstream task.}

\subsection{Problem Definition}
% The logic-level designs stored in BENCH format are converted to AIG graphs first proposed in \cite{brayton1990multilevel} for logic circuites representations. AIGs are directed acyclic graph (DAG) \cite{mishchenko2006dag} representation with 2-input AND function (one node type) and NOT or buffer function (two edge types)  \cite{amaru2015majority}.
The logic-level designs stored in BENCH format are first converted into AIGs, a format initially proposed in \cite{brayton1990multilevel} for representing logic circuits. An AIG is a type of Directed Acyclic Graph (DAG) \cite{mishchenko2006dag}, characterized by nodes representing 2-input AND functions and edges signifying either NOT or buffer functions \cite{amaru2015majority}. This structure simplifies the representation and manipulation of boolean functions, facilitating various logic synthesis and optimization tasks.
% \FF{The logic-level designs stored in BENCH format are converted to And-Inverted-Graphs (AIGs), first proposed in \cite{brayton1990multilevel}, for representing logic circuits. AIGs are directed acyclic graphs (DAGs) \cite{mishchenko2006dag} that use 2-input AND gates as nodes and NOT or buffer functions as edge types \cite{amaru2015majority}.}

Consequently, the input AIGs are transformed into attributed directed acyclic graphs (DAGs) $\mathcal{G}=(\mathcal{V},\mathcal{E},\mathcal{X}^{v})$, comprising $N$ nodes and $E=|\mathcal{E}|$ directed edges. Node features $\mathcal{X}^{v}$ are categorized into two types: $\mathcal{X}^{v}_{type}\in \{0,1,2\}$ and $\mathcal{X}^{v}_{inverted}\in \{0,1,2\}$. These features represent the classification of nodes (input, output, or intermediate) and the count of inverted predecessors for each gate, respectively. The set of recipe, denoted as $\mathcal{R} = \{r_{1}, r_{2}, ..., r_{R}\}$, comprises $R$ recipes, each containing $M$ sequentially ordered heuristics, selected from a set $\mathcal{T} = \{t_{1}, t_{2}, \ldots, t_{C}\}$, which consists of $C$ types of heuristics. Each element in $\mathcal{T}$ is an optimizer.
% \MB{pick something more suitable transformations/operations/etc}.

% \MB{what is the formulation for the model?}

% \MB{what is the formulation for the downstream task?}

% \cite{amaru2015majority} AIG  directed acyclic graph (DAG) representation with 2-input AND function (nodes) and NOT function (dotted edges)

% \cite{mishchenko2006dag} DAG representation. and rerwriting optimization command

% \cite{brayton1990multilevel} the first and principal paper for AIG and logic circuites representations.

%%%%%%%%%%%%%%%%%%%%%%%%%%%%%%%%%%%%%%%%%%%%%%%%%%%%

%uncomment end raika

%%%%%%%%%%%%%%%%%%%%%%%%%%%%%Task Definition%%%%%%%%%%%%%%%%%%%%%%%%%%%%
\subsection{Task Definition}  
% It is the process of converting a network of technology-independent logic gates into a network comprising logic cells on the Layout of target IC (such as FPGA).
\textbf{Technology Mapping:} The process involves transforming a network of technology-independent logic gates into a network consisting of logic cells tailored to the layout of a target integrated circuit (IC), such as a Field-Programmable Gate Array (FPGA). The metrics such as delay, area, space can be measured after this step. 

% \MB{this goes into the experimental setup. You just assume we have the dataset and is represented by some notation. Data generation is experimental details.}
% \MB{\RK
\textbf{Ground Truth Generation:} In the EDA workflow, we can optimize logic-level circuites modeled by AIG graphs $\mathcal{G}$ given the recipe through the time consuming tools like ABC. After sequentially applying the \( i^{th} \) heuristic and optimizing the logic-level representation, technology mapping is then applied to map the optimized version, with the ground truth QoR after each step shown by \( y^{i} \). The final QoR after $M^{th}$ heuristics, named $y$, is the ground truth of the model that our proposed model is expected to predict given the input AIGs and recipe during inference mode.
% }

\subsubsection{Self-Supervised Learning Auxiliary task}
As an auxillary task helping the model during training, we train the model to predict the trajectory of QoR evolution. To achieve this, the model step by step predicts all $M$ intermediate QoRs in a causal manner. For predicting the $i^{th}$ QoR, the model takes as input all heuristics before the $i^{th}$ step, $(r_{1},r_{2},..., r_{i})$. 
This predictive SSL task by it self is aligned with architecture of our transfomrer decoder to help model benefits from extra information within intermediate steps. 
% \FF{As an auxiliary task to assist the model during training, we train it to predict the trajectory of QoR evolution. To this end, the model step-by-step predicts all \(M\) intermediate QoRs in a causal manner. Specifically, to predict the \(i^{th}\) QoR, the model is given all heuristics up to the \(i^{th}\) step, \((r_{1}, r_{2}, \ldots, r_{i-1})\). This predictive SSL task aligns with the architecture of our transformer decoder, helping the model benefit from the additional information within intermediate steps.
% }
% In this section, we propose QoR trajectory prediction task utilized during training to empower Graph Encoders, Transformer Decoder as well as Recipe Encoders.
% \RK{mention  losses and the task}

\subsubsection{Final QoR prediction}
The QoR prediction task aims to find the learnable function $f_\gamma$ that maps a pair of recipe and AIG graph, $(\mathcal{G}, r_{i})$ to a numerical value $\hat{y}$ which is a prediction of final ground truth QoR such as delay and area.  More specifically, the task is as follows:

\begin{align}
    \hat{y} = f_\gamma(\mathcal{G}, r_{i})
\end{align}

The final loss which is Mean Squared Error is defined as follows:

\begin{align}
    loss = \sum_{k=1}^{M}MSE(\hat{y}_{k}, y_{k})
\end{align}
Where  $y_{k}$ represents the ground truth QoR of $k^{th}$ step.

\subsection{Architecture}
The architecture of our system consists of three main components: the graph encoder, the recipe encoder, and the fusion mechanism coupled with the decoder. The novelty of our work primarily lies in the enhancements to the decoding process as well as modifications to the other components. These innovations are detailed below, emphasizing the unique aspects of our approach within each block.

\subsubsection{Graph Encoder}
% In the AIG side, we use graph encoder $f_{\theta}$, according to the following formmulation:
In the AIG segment, we employ a graph encoder  $\mathcal{E}_{\theta}$ based on the following formulation:

\begin{align}
    \boldsymbol{H} = \mathcal{E}_{\theta}(\mathcal{G})\,
    % ,\quad \quad \quad
                 % \boldsymbol{Z} = \textsc{Decoder}_\gamma(\boldsymbol{H})\,,             
\end{align}

Here, $\boldsymbol{H} =\{h_{1}, h_{2}, ... ,h_{N}\}$ represents the set of node embeddings and $h_{i} \in \mathbb{R}^{d_{h}}$ is $i^{th}$ node embedding. Similar to the approach used in OpenABC \cite{chowdhury2021openabc}, we utilize a Graph Convolutional Neural Network (GCN) with the same node embedding approach to encode the AIGs.

\subsubsection{Level-wise Node Pooling}
Although the GCNs can encode graphs properly, they are unable to particularly incorporate the inductive bias of DAGs. To benefit from this bias to improve the graph representation learner, we propose an unique pooling mechanism called level-wise pooling.
% \FF{Although GCNs can encode graphs effectively, they are unable to specifically incorporate the inductive bias of DAGs. To leverage this bias and improve the representation learning, we propose a unique pooling mechanism called level-wise pooling.
% }

Similar to the work proposed in \cite{luo2024transformers} the depth of a node $v$ is calculated through the function $depth(v)$ because of the partial order intrinsic to the DAG. The set of node embeddings can be partitioned based on the depth of the each node as follows:
% \FF{Similar to the work proposed in \cite{luo2024transformers}, the depth of the nodes is calculated through the function \(depth(v)\) due to the partial order intrinsic to the DAG. The set of node embeddings can be partitioned based on the depth of each node as follows:}
 
\begin{align}
    % \boldsymbol{H} = f_{\theta}(\mathcal{G})\,
    \boldsymbol{H} = \{ \boldsymbol{H}^{0}, \boldsymbol{H}^{1}, ... , \boldsymbol{H}^{D} \}
    % ,\quad \quad \quad
                 % \boldsymbol{Z} = \textsc{Decoder}_\gamma(\boldsymbol{H})\,,
\end{align}

Where $\boldsymbol{H}^{l}$ represents the set of embeddings for nodes located at $l^{th}$ level of AIG. 
Finally, the AIG graph is represented as a sequence for the decoder layer described in the next section. The sequence is derived according to the combination of mean and max poolings as follows:

\begin{align}
    % \boldsymbol{H} = f_{\theta}(\mathcal{G})\,
    \bar{H} = (\bar{h}_{0}, \bar{h}_{1}, ...,\bar{h}_{D}), \quad \quad \quad
    \bar{h}_{l}  = \textsc{POOL}_{\text{level}}(\boldsymbol{H}^{l}) 
    % ,\quad \quad \quad
                 % \boldsymbol{Z} = \textsc{Decoder}_\gamma(\boldsymbol{H})\,,
\end{align}

where $\bar{H}$ is the sequence of representation of DAG and $\bar{h}_{l} \in \mathbb{R}^{2\times d_{h}}$ is the embedding of $l^{th}$ level of AIG. The $\textsc{POOL}_{\text{level}}$ is the concatenation of mean and max poolings.

% $\boldsymbol{H}^{l} = \{h_{l_{1}}, h_{l_{2}}, ... ,h_{l_{N}}\}$ located at $l^{th}$ depth is shown by $h_{l,}$ derived from 
% Although the GCNs can encode graphs properly, they are unable to particularly incorporate the inductive bias of DAG. To benefit from this bias to improve the representation learner, we propose an unique pooling mechansism called level-wise pooling. In this pooling mechanism, we perform pooling across all nodes within the same level of DAG. In other words, all nodes comming form the same level are mean and max pooled.

\begin{table*}
    \caption{Performance comparison of delay and area prediction models across datasets, evaluated using MAPE (Avg. ± Std.) (\%).}
    \centering
    \renewcommand{\arraystretch}{1.2}
    \begin{adjustbox}{width=\textwidth}
    \begin{tabular}{|c|c|c|c|c|c|c|c|c|c|c|c|c|}
        \hline
        \multirow{3}{2.5cm}{\textbf{Architecture}} & \multicolumn{6}{c|}{\textbf{IP-Inductive}} & \multicolumn{6}{c|}{\textbf{Recipe-Inductive}} \\
        \cline{2-13}
        & \multicolumn{2}{c|}{\textbf{EPFL}} & \multicolumn{2}{c|}{\textbf{OABCD}} & \multicolumn{2}{c|}{\textbf{PD}} & \multicolumn{2}{c|}{\textbf{EPFL}} & \multicolumn{2}{c|}{\textbf{OABCD}} & \multicolumn{2}{c|}{\textbf{PD}} \\
        \cline{2-13}
        & delay & area & delay & area & delay & area & delay & area & delay & area & delay & area \\
        \hline
        \midrule
        OpenABC (Baseline) & 4.18 $\pm$ 0.07 & 2.46 $\pm$ 0.03 & 23.66 $\pm$ 0.19 & 2.71 $\pm$ 0.02 & 15.88 $\pm$ 0.41 & 3.57 $\pm$ 0.10 & 5.35 $\pm$ 0.04  & 2.74 $\pm$ 0.01  & 16.79 $\pm$ 0.08  & 2.72 $\pm$ 0.01  & 16.69 $\pm$ 0.4  & 2.04 $\pm$ 0.01  \\
        LOSTIN  (GIN + LSTM) \cite{wu2022lostin} & 3.96 $\pm$ 0.02 & 2.30 $\pm$ 0.01 & 24.61 $\pm$ 0.03 & 2.35 $\pm$ 0.07 & 16.76 $\pm$ 0.13 & 3.46 $\pm$ 0.17 & 5.14 $\pm$ 0.04 & 2.73 $\pm$ 0.04 & 17.47 $\pm$ 0.04 & 1.77 $\pm$ 0.03 & 17.61 $\pm$ 0.07 & 2.03 $\pm$ 0.08 \\
         (GraphSage + Transformer) \cite{yang2022prediction} & 3.96 $\pm$ 0.02 & 2.39 ± 0.00 & 24.58 $\pm$ 0.13  & 3.77 $\pm$ 0.00  & 17.09 $\pm$ 0.08  & 3.81 $\pm$ 0.05 & 5.79 $\pm$ 0.02 & 2.96 $\pm$ 0.00 & 18.33 $\pm$ 0.60 & 3.4 $\pm$ 0.06 & 18.04 $\pm$ 0.02 & 2.43 $\pm$ 0.04 \\
         GNN-H \cite{wu2022ai} & 3.96 $\pm$ 0.03 & 2.34 $\pm$ 0.02  & 24.31 $\pm$ 0.27 & 2.33 $\pm$ 0.06 & 16.80 $\pm$ 0.06 & 3.59 $\pm$ 0.12 & 5.79 $\pm$ 0.01 & 2.61 $\pm$ 0.06 & 16.41 $\pm$ 0.40 & 1.58 $\pm$ 0.02 & 17.52 $\pm$ 0.05& 2.15 $\pm$ 0.01\\\hline
        LSOformer (ours) &\textbf{3.94 $\pm$ 0.02 } &\textbf{2.24 $\pm$ 0.02} & \textbf{22.63 $\pm$ 0.04} & \textbf{2.34 $\pm$ 0.05}  & \textbf{13.17 $\pm$ 0.28 } & \textbf{3.43 $\pm$ 0.05 } & \textbf{4.74 $\pm$ 0.01} & \textbf{2.45 $\pm$ 0.03} & \textbf{16.39 $\pm$ 0.16} & \textbf{1.21 $\pm$ 0.03} & \textbf{14.55 $\pm$ 0.02} & \textbf{1.92 $\pm$ 0.03} \\
        Improvement \% vs Baseline & 5.74\%$\uparrow$ & 8.94\%$\uparrow$ & 4.35\%$\uparrow$ & 13.65\%$\uparrow$ & 17.06\%$\uparrow$ & 3.92\%$\uparrow$ & 11.40\%$\uparrow$ & 10.58\%$\uparrow$  & 2.38\%$\uparrow$ & 55.5\%$\uparrow$  & 12.82\%$\uparrow$ & 5.88\%$\uparrow$ \\
        \bottomrule
    \end{tabular}
    \end{adjustbox}
    \label{tab:maintable}
\end{table*}
\subsubsection{Heuristic Tokenization and Embedding}

Each recipe, denoted as \( r_i = (u_1, u_2, \ldots, u_M) \), where each \( u_j \in \mathcal{T} \) represents the \( j \)-th heuristic within the recipe, is tokenized using a one-hot vector of length \( C \). A lookup table is utilized to generate learned embeddings to convert the input tokens to vectors $\bar{u}_{i}~\in~\mathbb{R}^{2\times d_{h}}$ within the sequence of embeddings $\bar{r}_{i} = (\bar{u}_{1}, \bar{u}_{2}, ..., \bar{u}_{M})$.

% \begin{align}
%     % \boldsymbol{H} = f_{\theta}(\mathcal{G})\,
%     Embedding(onehot(r_{i})
%     \bar{H} = (\bar{h}^{0}, \bar{h}^{1}, ...,\bar{h}^{D}), \quad \quad \quad
%     \bar{h}^{l}  = pooling(\boldsymbol{H}^{l}) 
%     % ,\quad \quad \quad
%                  % \boldsymbol{Z} = \textsc{Decoder}_\gamma(\boldsymbol{H})\,,
% \end{align}

% \MB{It is not clear how the two hidden embeddings out of the two pathways are fused using cross attention and the decoder-only transformer.}\RK

% - 1. define in formal language the entire representation learning pipeline for the task; the embedding layer, the encoding layer (two information processing pathways), the decoding layer where we formulate as the novel contribution to predict intermediate values.
% - - provide the details for the decoder-only transformer (both training and inference) where we map the input sequence to the output sequence.

\subsubsection{Transformer Decoder}
Once the recipes are embedded and the AIG is converted to a sequence, inspired by Machine Translation models \cite{vaswani2017attention}, we propose a sequence to sequence alignment module using only transformer decoder module to fuse the AIG sequence with heuristics' embeddings. In other words, this transformer module maps AIG sequence and heuristics' embeddings to the trajectory of QoRs in causal manner. More specifically: 

\begin{equation}
    \hat{Y} = \texttt{Transformer}(\bar{H}, \bar{r}_{i})
\end{equation}
\begin{equation}
    \hat{Y} = (\hat{y}_{1},...,\hat{y}_{M})
\end{equation}
% \begin{align}
%     \hat{Y} = \texttt{Transformer}(\bar{H}, \bar{r}_{i}),\quad\quad\quad\\
%     \hat{Y} = (\hat{y}^{1},...,\hat{y}^{M})
% \end{align}

where $\hat{Y}$ consists of the QoRs of the trajectory. Our transformer decoder containing 6 main components is designed and  formulated as follows:

% \MB{why do we need to talk about PE?}\RK
\textbf{Positional Encoding:}
Sine/Cosine positional encoding ($PE$) is added to the sequence of heuristics embeddings element-wise as follows:
%uncomment start raika
\begin{align}
    PE_{(m,2k)} = \sin\left(\frac{m}{10000^{k/d_{h}}}\right)\\
    PE_{(m,2k+1)} = \cos\left(\frac{m}{10000^{k/d_{h}}}\right)
\end{align}
%% uncomment end raika
\begin{align}
    z_i = \bar{r}_{i} + PE
\end{align}
% \FF{Please explain what $z_i$ is} 
Where 
% $m$ is the position index in the sequence,  k is dimension index, and
$z_i$ is the sequence of positionally encoded embeddings of Optimizers.

\textbf{Masked Multi-Head Self-Attention:} If $Q = z_iW_q \in \mathbb{R}^{M \times 2d_{h}}$, $K = z_iW_k \in \mathbb{R}^{M \times 2d_{h}}$, and $V = z_iW_v \in \mathbb{R}^{M \times 2d_{h}}$ are the query, key, and value matrices, the contextualized heuristics' embedding ($\bar{z_i}$) is derived as follows:
\begin{align}
    \bar{z_i} = Attention(Q, K, V) = softmax\left(\frac{QK^T}{\sqrt{d_k}} + \bar{M}\right)V
\end{align}
The mask $\bar{M}$ is typically an upper triangular matrix with zeros on and below the diagonal and \(-\infty\) above the diagonal for each sequence in the batch. This structure effectively imposes causality and ensures that each position in the sequence can only attend to itself and previous positions, not to any future positions.

\textbf{Cross-Attention:}
This module computes  the attention weights between the element of AIG sequence and contextualized heuristics' embeddings. Finally, a Feed-Forward layer (FFN) with Relu function is applied element-wise to the sequence.
\begin{align}
    \widetilde{h}^{i} = \text{FFN}(\text{MultiHeadAttention}(\bar{z_i}, \bar{H}, \bar{H}))
\end{align}
% \textbf{Feed-Forward Network:}
% Feed-forward networks remain unchanged in the causal setup:
% \begin{align}
%     \text{FFN}(x) = \max(0, xW_1 + b_1)W_2 + b_2
% \end{align}
% \FF{why $\bar{H}$ is repeated in the formula?}\RK
Where $\widetilde{h}^{i} = (\widetilde{u}_{1},\widetilde{u}_{2},...,\widetilde{u}_{M})$ is the output sequence of embeddings for input AIG and $i^{th}$ recipe; $\widetilde{u}_{i} \in \mathbb{R}^{2 \times d_{h}}$  represents generated embedding correspoding to $i^{th}$ QoR.

%% uncomment start raika
\textbf{Add \& Norm:} Each sub-layer (self-attention and feed-forward network) includes a residual connection followed by layer normalization:
\begin{align}
    \text{LayerNorm}(x + \text{Sublayer}(x))
\end{align}
%% uncomment end raika
% \FF{what is $x$}

\textbf{MLP Regressor:}
Once the embeddings for each QoR has been generated sequentially, we use a MLP module to map the embeddings to final QoR as follows:

\begin{align}
    \hat{y}_{i} = MLP(\widetilde{h}^{i})
\end{align}
% \vspace{-4pt}
% \subsubsection{Open ABC Dataset}
\begin{table}[b]
\centering
\caption{Statistics for benchmark datasets.}
\resizebox{0.6\linewidth}{!}{%
% \resizebox{0.80\linewidth}{!}{%
\begin{tabular}{c|ccc}
\hline 
 Dataset & EPFL & OABCD  & PD  \\\hline
\midrule
Number of Circuits  & 15 & 29 &   118 \\ 
Avg \# Nodes & 7629.25 & 12092.25 &  21746.99  \\ 
Avg \# Edges & 31035.46  & 63640.44 &  40436.10  \\ 
Avg Depth & 1835.8 & 65.55 & 56.19     \\ 
Avg PI & 224.75 & 3118.96 &  1095.41   \\ 
Avg PO & 143.05 & 2803.04 &  867.05  \\  
% LastFM  & 1980  & 1,293,103  & 154,993  &  &  & \checkmark  & 68\% \tabularnewline \hline
% UCI  & 1899  & 59,835  & 13838  &  &  & \checkmark  & 62\% \tabularnewline \hline
% Enron  & 184  & 125,235  & 2215  &  &  &  & 92\% \tabularnewline \hline
% SocialEvolution  & 74  & 2,099,519  & 2506  &  &  &  & 97\% \tabularnewline \hline
\bottomrule
\end{tabular}%
}
\label{tab:stats} 
\end{table}

\section{Experimental Evaluation}
% \MB{call it "Experimental Evaluation"}

% \MB{you need some opening paragraph to stage what we are going to talk about in the following.}
% In this part, we designed experiments to strongly support three main contributions mentioned in this work. Then, we evaluate the performance of the proposed architecture in different setups and scenarios.

In this section, we have structured experiments to robustly validate the three main contributions outlined in this work. Following the experimental design, we assess the performance of the proposed architecture across various setups and scenarios, providing a comprehensive evaluation of its effectiveness and adaptability in handling different computational challenges. 
%%%%%%%%%%%%%%%%%%%%%%%%%%%%%%%%%%%%%%%%%%%%%%%%%%%%%%%%%%%%%%
% \MB{change it as follows:}
% \MB{- experimental setup
% - - datasets: what are the datasets we benchmark? some stats of space allows
% - - baselines: Among those introduces in the related work, which ones are comparable to ours? briefly present them.
% - - evaluation protocols/metrics: what are the metrics to measure performance? 
% }

\subsection{Tools}
% We use OpenROAD v1.0 [15] EDA to perform logic synthesis; it uses Yosys [16] as the frontend engine (currently v0.9). Yosys performs logic synthesis in conjunction with ABC [11]. It can
% generate a logic minimized netlist for a desired QoR. We use networkx v2.6 for graph processing.
% For ML frameworks on graph structured data, we use pytorch v1.9 and pytorch-geometric v1.7.0.
% We collect area and timing of the AIG post-technology mapping using NanGate 45nm technology

% Yosys tool

% \cite{brayton2010abc} it is used in the stateof-the-art open source logic synthesis tool, ABC.

% \cite{ajayi2019openroad} OpenRoad tool 

% \cite{wolf2016yosys} yosys tool

Commercial EDA tools offer standardized operating systems [5], whereas the academic, open-source tool ABC \cite{brayton2010abc} provides specialized heuristic commands for circuit optimization.
We employ OpenROAD v1.0  EDA  \cite{ajayi2019openroad} for logic synthesis, with Yosys \cite{wolf2016yosys}, version 0.9, serving as the frontend engine. Yosys works in collaboration with ABC to carry out post-technology mapping and generate a minimized logic circuit specifically optimized for the final QoR.
Post-technology mapping, the area and delay of these designs are evaluated using the NanGate 45nm technology library and the 5K\_heavy wireload model. Furthermore, we utilize PyTorch v1.13 and PyTorch Geometric v2.2.0 to generate AIGs.

\subsection{Datasets}

Three datasets were employed for the experiments, each comprising a collection of circuit designs beside a set of recipe.  Since each dataset is used in different studies, we chose not to combine them to simplify comparisons with baseline methods. Additionally, we report performance metrics individually for each dataset. Logic-level circuit designs in the datasets are referred to as Intellectual Property (IP). A consistent set of 1500 unique recipes borrowed from \cite{chowdhury2021openabc} was applied across all datasets. Each recipe was constructed utilizing seven primary heuristics and their associated flags from the ABC toolkit. These heuristics include Balance, Rewrite (rw, rw -z), Refactor (rf, rf -z), and Re-substitution (rs, rs -z).

% \MB{make these textbf rather than subsubsection to save space}

\textbf{EPFL:} Introduced in 2015, the EPFL Combinational Benchmark Suite comprises combinational circuits specifically designed to test the capabilities of modern logic optimization tools \cite{amaru2015epfl}. This benchmark suite has been complemented by an open-source leaderboard, which aims to establish a new comparative standard within the logic optimization and synthesis community. The suite is categorized into arithmetic, random/control, and MtM circuits, with each circuit available in multiple formats including Verilog, VHDL, BLIF, and AIG.

\textbf{Open ABC Dataset (OABCD):} 
% This dataset consist of 29 IPs followed by a LSO pipeline.  The pipeline \cite{chowdhury2021openabc} proposes an end-to-end preprocessing pipeline to preprocess and run three regression-based downstream task depending on the stage of post-tech mapping are experimented. Moreover, in this baseline pipeline, there is an additional classification task performed to measure the effectiveness of the graph encoder for distinguishing graphs constructed from different IPs. 
This dataset comprises 29 circuits followed by a LSO pipeline. The pipeline, as proposed in \cite{chowdhury2021openabc}, features an end-to-end preprocessing system designed to handle and execute three regression-based downstream tasks, each dependent on the stage of post-technology mapping. Additionally, the baseline pipeline includes a classification task aimed at evaluating the efficacy of the graph encoder in distinguishing between graphs derived from different Circuits.

\textbf{Proprietary Dataset (PD):} This dataset includes 118 internal circuits, with the number of nodes per circuit approximately varying from 80 to 40,000. 
Additionally, detailed statistics for the number of nodes, depth, number of input nodes (PI), and number of output nodes (PO) in all datasets are provided in Table~\ref{tab:stats}.

%%%uncomment raika

\begin{table}[t]
    \centering
    \captionof{table}{Ablation study on Recipe Encoder (Recipe Enc.) and AIG Encoder (AIG Enc.) using the Freeze Probing setup, evaluated with MAPE. }
    \label{table:table3}
    \resizebox{0.98\linewidth}{!}{%
        \begin{tabular}{C{3cm} C{1.5cm} C{1.5cm} C{1.5cm} C{1.5cm} C{1.5cm} C{1.5cm}}
            \toprule
             Pre-text task setup & QoR Prediction &  Frozen GE & Frozen RE & EPFL & OABCD  \\ \hline
            \midrule
             Random Freeze & \xmark & \cmark & \cmark & 5.05 & 26.74 \\
             Random AIG Enc. & \cmark & \cmark & \xmark & 4.32 & 24.37 \\
             Random recipe Enc. & \cmark & \xmark & \cmark &  4.51 & 26.13 \\
             \midrule
             Freezed LSOformer & \cmark & \xmark & \xmark &  3.98 & 23.5\\
             % & \xmark & \xmark & $$ & $\mathbf{}$ & $$\\
             \midrule
             Supervised baseline & \xmark & \xmark & \xmark & 4.26 & 23.58\\
             Supervised LSOformer & \xmark & \xmark & \xmark & 3.94 & 22.58\\
            % \midrule
            % \multirow{4}{*}{LastFM} & \xmark & \xmark & \xmark & $0.960$ \\
            % & \cmark & \xmark & \xmark & $0.961$ \\
            % & \cmark & \cmark & \xmark & $0.965$ \\
            % & \cmark & \cmark & \cmark & $\mathbf{0.976}$ \\
            % \midrule
            % \multirow{4}{*}{UCI} & \xmark & \xmark & \xmark & $0.981$ \\
            % & \cmark & \xmark & \xmark & $0.983$ \\
            % & \cmark & \cmark & \xmark & $0.987$ \\
            % & \cmark & \cmark & \cmark & $\mathbf{0.993}$ \\
            % \midrule
            % \multirow{4}{*}{SocialEvolution} & \xmark & \xmark & \xmark & $0.987$ \\
            % & \cmark & \xmark & \xmark &  $0.987$\\
            % & \cmark & \cmark & \xmark & $0.989$ \\
            % & \cmark & \cmark & \cmark & $\mathbf{0.991}$ \\ \hline
            \bottomrule
        \end{tabular}%
        }
\end{table}

 \begin{table}[b]
    \centering
    \captionof{table}{Ablation study on the decoder and SSL auxiliary task, evaluated with the MAPE metric. P.Q.T. stands for Predicting QoR trajectory.}
    \label{table:table4}
    \resizebox{0.95\linewidth}{!}{%
        % \begin{tabular}{CCC|CC }
        \begin{tabular}{C{3cm} C{1.5cm} C{1.5cm}|C{1.5cm} C{1.5cm}  }
            \toprule
             Setup name & P.Q.T. & Transformer decoder & EPFL & OABCD  \\ \hline
            \midrule
            OpenABC (Baseline) &\xmark & \xmark & 4.26 & 23.58  \\
            OpenABC + SSL &\cmark & \xmark & 4.16 & 26.48 \\
            Transformer Decoder &\xmark & \cmark & 4.01 & 24.62 \\
            LSOformer&\cmark & \cmark & \textbf{3.94} & \textbf{22.58} \\
            %  Random Freeze & \xmark & \cmark & \cmark & $$ & $$ & $$ \\
            %  Random Graph Encoder & \cmark & \cmark & \xmark & $$ & $$ & $$ \\
            %  Random Recipe Encoder & \cmark & \xmark & \cmark & $$ & $$ & $$ \\
            %  \midrule
            %  LSOformer & \cmark & \xmark & \xmark & $$ & $\mathbf{}$ & $$\\
            %  % & \xmark & \xmark & $$ & $\mathbf{}$ & $$\\
            %  \midrule
            %  Supervised & \xmark & \xmark & \xmark & $$ & $\mathbf{}$ & $$\\
            % % \midrule
            % % \multirow{4}{*}{LastFM} & \xmark & \xmark & \xmark & $0.960$ \\
            % % & \cmark & \xmark & \xmark & $0.961$ \\
            % % & \cmark & \cmark & \xmark & $0.965$ \\
            % % & \cmark & \cmark & \cmark & $\mathbf{0.976}$ \\
            % % \midrule
            % % \multirow{4}{*}{UCI} & \xmark & \xmark & \xmark & $0.981$ \\
            % % & \cmark & \xmark & \xmark & $0.983$ \\
            % % & \cmark & \cmark & \xmark & $0.987$ \\
            % % & \cmark & \cmark & \cmark & $\mathbf{0.993}$ \\
            % % \midrule
            % % \multirow{4}{*}{SocialEvolution} & \xmark & \xmark & \xmark & $0.987$ \\
            % % & \cmark & \xmark & \xmark &  $0.987$\\
            % % & \cmark & \cmark & \xmark & $0.989$ \\
            % % & \cmark & \cmark & \cmark & $\mathbf{0.991}$ \\ \hline
            \bottomrule
        \end{tabular}%
        }
\end{table}

%%%%%%%%%%%%%%%%%%%%%%%%%%%%%%%%%%%%%%%%%%%%%%%%%%%%%%%%%%%%%%
\subsection{Experimental Protocol}
The efficacy of our model is assessed under both transductive and inductive settings to demonstrate its superiority over competing architectures. 
The dataset is split into training and validation sets with ratios of 0.66 and 0.33, respectively, for both setups. Additionally, zero-mean normalization is applied to normalize the final QoRs. Additionally, intermediate QoRs are normalized using the mean and standard deviation derived from the final QoRs. 
In the ablation study, we focus exclusively on the inductive setting, which presents the most challenging scenario for the LSO problem. In these experiments, given the challenges associated with predicting timing, delay are selected as the target variable for QoR prediction.

\textbf{Inductive Setup (IP-Inductive):} In this setup, the circuits used during the training phase are different from those employed in testing. It is noteworthy that all recipes from the sample set are observed and applied to the circuits introduced to the model during training.
% In this setup, the IPs seen during training is different from  IPs used in testing. It's worth mentioning that all the OS from sample set is seen and pplied to the IPs fed to the model during training. 

\textbf{Transductive Setup (Recipe-Inductive):} In the transductive setup, each circuit is exposed during training, but separate recipes are used for testing. Specifically, the set of recipes is partitioned into test and validation subsets, and all Circuits are visible during both training and testing phases, albeit with different recipes.

% In transductive setup, each IP is seen during training but separated OSs is used for testing. In other words, the set of OSs is partioned into test and validation and all IPs is seen during both of training and testing but with different OSs. 

%%%%%%%%%%%%%%%%%%%%%%%%%%%%%%%%%%%%%%%%%%%%%%%%%%%%%%%%%%%%%%
\subsubsection{Implementation Details and Parameter Settings}

%%%%%%%%%%%%%%%%%%%%%%%%%%%%%%%%%%%%%%%%%%%%%%%%%%%%%%%%%%%%%%
In accordance with \cite{chowdhury2021openabc}, for delay prediction, the AIG encoder uses two layers of GCN with a hidden embedding size of 32, while for area prediction, it employs 10 layers of GIN with batch normalization. Additionally, mean and max pooling were applied combined to generate embeddings for each level of AIG; hence, the dimension of the transformer decoder is set to 64.
The number of levels is determined by the maximum depth of existing circuits for each dataset. Circuits with a lower depth are zero-padded to align their length. Early stoping has been utilized to report the best validation result in Table~\ref{tab:maintable}.
\subsection{Results}

\begin{table}[t]
    \centering
    \captionof{table}{Ablation study on different decoder layers for autoregressive and causal QoR trajectory prediction. Values represents MAPE metric.}
    \label{table:table5}
    \resizebox{0.8\linewidth}{!}{%
        \begin{tabular}{ C{3cm} C{1.5cm}  C{1.5cm} C{1.5cm}}
            \toprule
             Decoder & EPFL &  OABCD & PD \\ \hline
            \midrule
            MLP  & 5.33 & 17.38 & 17.75\\
            MLP + multi-task & 5.15 & 16.85 & 15.57\\
            Auto-regressive LSTM & 5.09 & 17.53 & 17.32\\
            Transformer & \textbf{4.73} & \textbf{16.57} & \textbf{14.65}\\
            % graph encoding  & $$ & $$ & $$ \\
            %  GCN & CNN & MLP & \cmark & $$ & $$ & $$ \\
            % \midrule
            %  GraphSage & Transformer & MLP & \xmark & $$ & $$ & $$ \\
            %  GraphSage & Transformer & MLP & \cmark & $\mathbf{s}$ & $$ & $$ \\
            % \midrule
            %  GIN & LSTM & MLP & \xmark & $\mathbf{}$ & $$ & $$ \\
            %  GIN & LSTM & MLP & \cmark & $\mathbf{}$ & $$ & $$ \\
            % \midrule
            %  GCN & Transformer & MLP & \xmark & $\mathbf{}$ & $$ & $$ \\
            %  GCN & Transformer & MLP & \cmark & $\mathbf{}$ & $$ & $$ \\
            %  GCN & Transformer & Transformer & \xmark & $\mathbf{}$ & $$ & $$ \\
            %  GCN & Transformer & Transformer & \cmark & $\mathbf{}$ & $$ & $$ \\
            % \midrule
            % \multirow{4}{*}{EPFL} & \xmark & \xmark & \xmark & $0.960$ \\
            % & \cmark & \xmark & \xmark & $0.961$ \\
            % & \cmark & \cmark & \xmark & $0.965$ \\
            % & \cmark & \cmark & \cmark & $\mathbf{0.976}$ \\
            % \midrule
            % \multirow{4}{*}{HS} & \xmark & \xmark & \xmark & $0.981$ \\
            % & \cmark & \xmark & \xmark & $0.983$ \\
            % & \cmark & \cmark & \xmark & $0.987$ \\
            % & \cmark & \cmark & \cmark & $\mathbf{0.993}$ \\
            \bottomrule
        \end{tabular}%
     }
     \vspace{-9pt}
\end{table}

\subsubsection{Comparison Study Result}
% \RK{this part may go to the appendix if we do not have space}
In Table \ref{tab:maintable}, we evaluate the MAPE performance of various QoR prediction models against the baseline in multi trials across two setups for predicting delay and area. The baseline model is the architecture proposed in the OABCD paper \cite{chowdhury2021openabc}. In the IP-inductive setup, the LSOformer surpasses all existing models and outperforms the baseline by 5.74\%, 4.35\%, and 17.06\% for the EPFL, OABCD, and PD datasets, respectively, in terms of delay prediction. For area prediction, LSOformer achieves a lower MAPE across all datasets compared to the baseline and nearly all models. However, on the OABCD dataset, the GNN-H model achieves the highest performance, demonstrating a 14\% improvement over the baseline. This pattern holds in the recipe-inductive setup for both delay and area predictions across the three datasets, where LSOformer exceeds the baseline by 11.40\%, 2.38\%, and 12.82\% for the EPFL, OABCD, and PD datasets in delay prediction, respectively.

\begin{figure*}[t]
    \centering
    % \resizebox{0.3\linewidth}{!}{
    \begin{subfigure}{1\columnwidth}
        \centering
        \includegraphics[width=\textwidth]{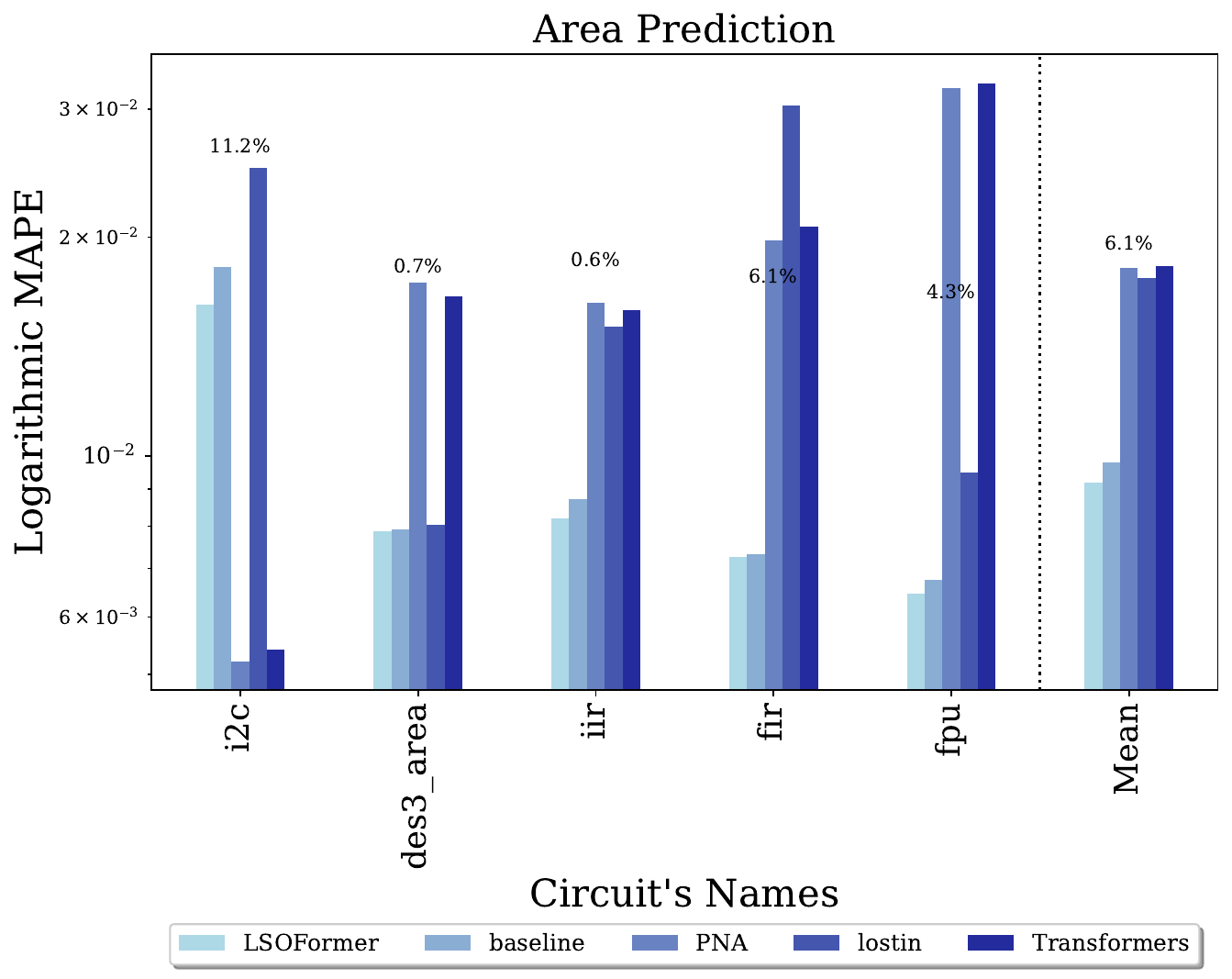} % specify the path to image1
        \caption{}
        \label{fig:sub1}
    \end{subfigure}
    % }
    \hfill
    % \resizebox{0.3\linewidth}{!}{
    \begin{subfigure}{1\columnwidth}
        \centering
        \includegraphics[width=\textwidth]{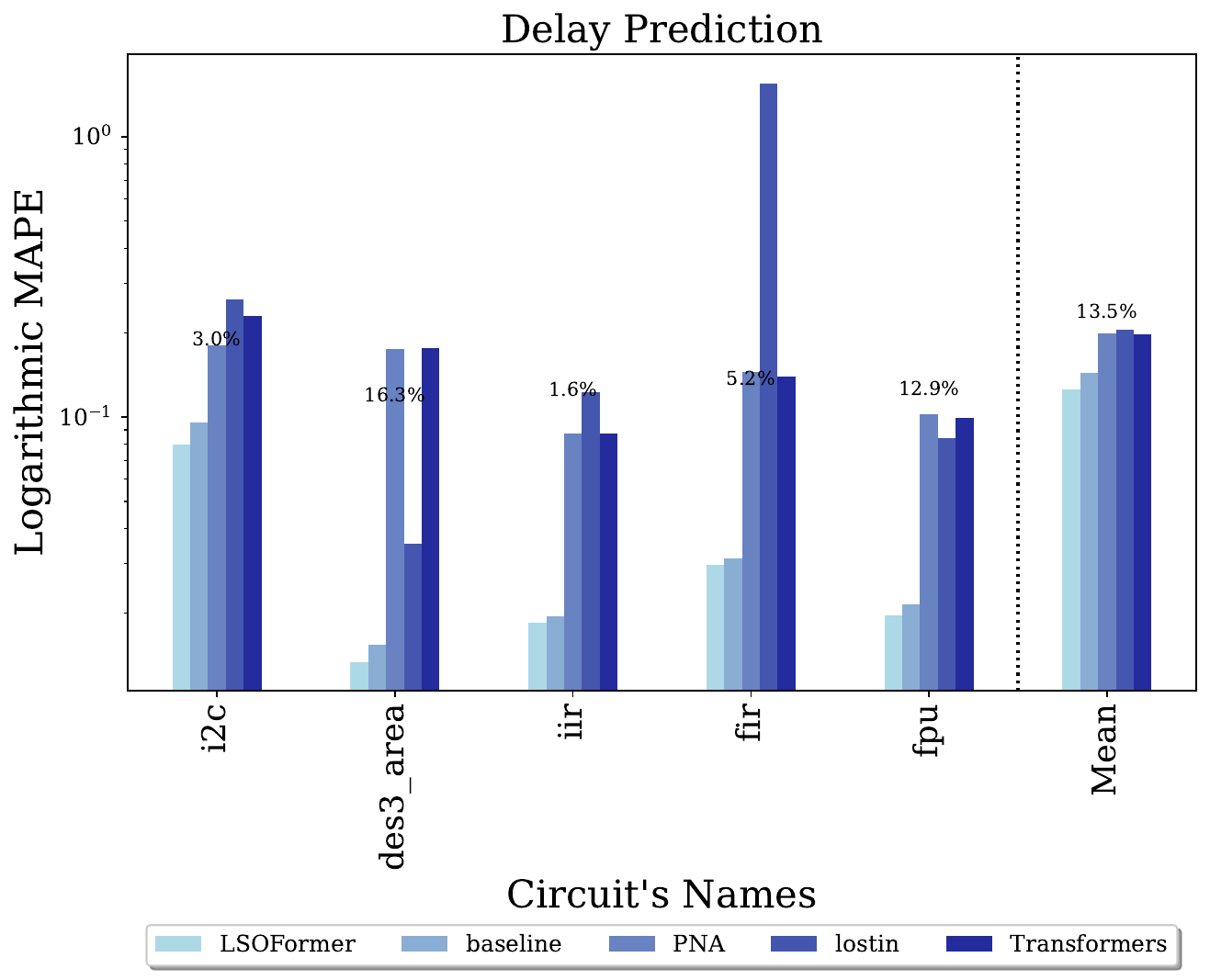} % specify the path to image2
        \caption{}
        \label{fig:sub2}
    \end{subfigure}
    % }
    \caption{Mean and circuit-wise performance comparison across test circuits for Area (a) and Delay (b).}
    \label{fig:fig2}
\end{figure*}

%%%%%%

% \begin{table}[htbp]
%     \centering
%     \caption{Example of a multi-column table}
%     \label{tab:multicol}
%     \begin{tabular}{|c|c|c|}
%         \hline
%         \multirow{2}{*}{\textbf{Header 1}} & \multicolumn{2}{c|}{\textbf{Header 2}} \\ \cline{2-3} 
%                                             & \textbf{Subheader 1} & \textbf{Subheader 2} \\ \hline
%         Row 1, Column 1 & Row 1, Column 2 & Row 1, Column 3 \\ \hline
%         Row 2, Column 1 & Row 2, Column 2 & Row 2, Column 3 \\ \hline
%     \end{tabular}
% \end{table}

\subsubsection{Ablation Study}
% \MB{have the following in this section}
% \MB{
% - ablation studies
% - - ablations studies to show the effectiveness of our contribution module/piece by module/piece
% - - - one ablation study is on the probing protocols (random freeze, pre-train and then fine-tune, and joint optimization)
% - - - one ablation study is on decoder-only transformer in contrast to MLP (one-shot, multi-shot), RNN, LSTM, etc.
% }

%%%%%%%%%%%%%%%%%%%%%%%%%%%%%%%%%%%%%%%%%%%%%%%%%%%%%%%%%%%%%%%
  We assess the impact of the SSL auxiliary loss integrated to the training pipeline in IP-inductive setup using MAPE as evaluation metric. Similar to the benchmark SSL works \cite{liu2022graph}, for the random setup, different parts of architecture are freezed to generate the lower bound of performance for SSL task. The difference between the results of random encoders and supervised models is considered the gap that the frozen model can potentially bridge in the best-case scenarios. Freezed LSOformer in Table~\ref{table:table3}, benefiting from QoR trajectory prediction, has performance near that of the supervised LSOformer and shows better results compared to the baseline, although it has not trained on final QoR.
 
%%%%%%%%%%%%%%%%%%%%%%%%%%%%%%%%%%%%%%%%%%%%%%%%%%%%%%%%%%%%%%%
Table~\ref{table:table4} details an ablation study on the integration of transformer architecture and SSL auxiliary tasks within the LSOformer model. It examines the impact of combining QoR trajectory predictions with various architectures. The results indicate that integrating our proposed architecture with the SSL task significantly improves performance over the baseline and other configurations. Conversely, removing the SSL component results in performance that does not meet the baseline levels. Additionally, using the baseline architecture for QoR trajectory prediction shows limited improvement on the EPFL dataset and a decline on the OABCD dataset, underscoring the critical role of the SSL task in enhancing the efficacy of our transformer architecture. This suggests that with ample data, the transformer decoder is particularly effective for the LSO task.
%%%%%%%%%%%%%%%%%%%%%%%%%%%%%%%%%%%%%%%%%%%%%%%%%%%%%%%%%%%%%%%

%%%%%%%%%%%%%%%%%%%%%%%%%%%%%%%%%%%%%%%%%%%%%%%%%%%%%%%%%%%%%%%
 % we ablated on embedding fusion and decoder part using different sequence decoders. The baseline, which is shared MLP in addition to concatenation of recipe and AIG embeddings for predicting all intermediate and final QORs,  falls behind all the counterparts. Moreover, separated MLP for predicting each QOR (from beginning to final) to resemble multi-task setup. This setup shows significant improvement on all datasets compared to baseline. We have also tried auto-regressive LSTM module to auto-regresively train and inference models to learn QORs to impose causality to LSTM. This model shows improvement compared to baseline in all datasets except OABCD. As can be seen in Table~\ref{table:table5}, LSOformer beats baseline as well as all other methods across three datasets justifying the proper choose of our architecture design.

 The effects of varying the embedding fusion and decoder configurations using different sequence decoders are explored in this work. 
The baseline model, which uses a shared MLP to predict all QoRs from initial to final, performs worse than alternative designs. Alternatively, separate MLPs are employed for each QoR, adopting a multi-task learning approach.
 This configuration significantly enhances performance across all datasets compared to the baseline. Additionally, we experimented with an auto-regressive LSTM module designed to train and infer models auto-regressively, thereby introducing causality to LSTM. This model demonstrates improvements over the baseline in all datasets except for OABCD. As can be seen in Table~\ref{table:table5}, the LSOformer surpasses both the baseline and all other methods across the three datasets, affirming the efficacy of our architectural design choices.

\subsubsection{In-depth Analysis}
 We assess the performance of our model on a test set in comparison to established baselines. In this configuration, the model is trained using PD datasets and primarily evaluated using OABCD and EPFL datasets. The analysis involves a randomized execution across a predetermined set of IPs, with results illustrated in Figure~\ref{fig:fig2}. Each column, representing an individual IP, displays the MAPE in logarithmic scale for area and delay for each model, as shown in Figure~\ref{fig:fig2}a and Figure\ref{fig:fig2}b, respectively. The IP names are arranged in ascending order from left to right based on the number of nodes. The final column presents the average MAPE across all IPs within the EPFL and OABCD datasets. Overall, LSOformer surpasses all competing models in this setup, affirming its dominance in the LSO task. Additionally, LSOformer demonstrates a notable improvement in area performance, approximately 6\%, across all IPs. For delay, although the improvement margin is narrower, there is a consistent enhancement observed across all selected IPs. The improvement gain of LSOformer, calculated as the difference from the baseline divided by the baseline performance, is displayed above each bar. For all sampled IPs, both LSOformer and the baseline outperform other methods, except for the i2c IP, where PNA and Transformer demonstrate exceptional performance in predicting area.

%uncomment raika

\section{Conclusion}
This  paper introduces an innovative solution to the Quality of Results (QoR) prediction for  Logic Synthesis Optimization (LSO) problem. A novel architecture is enhanced and integrated with a Self-Supervised Auxiliary task, designed to forecast the intermediate metrics of designs. These metrics are equivalent to those generated following post-technology mapping at each optimization phase. Additionally, we introduce a distinctive pooling mechanism for encoding And-Inverter Graphs (AIGs), which is combined with heuristic inputs using a transformer decoder architecture to predict QoRs at each step in a causal manner. The efficacy of this architecture is validated against conventional models across three primary datasets. Additionally, we demonstrate the robust alignment of the proposed architecture with the Self-Supervised Learning (SSL) auxiliary task specifically defined for this problem.

\newpage
% \begin{verbatim}
%   \bibliographystyle{ACM-Reference-Format}
%   \bibliography{bibfile}
% \end{verbatim}
% \begin{verbatim}
%   \citestyle{acmauthoryear}
% \end{verbatim}
\bibliographystyle{ACM-Reference-Format}
\bibliography{references}

% \bibliography{references.bib}

\end{document}